\documentclass[runningheads]{llncs}
\usepackage[T1]{fontenc}
\usepackage{subcaption}   
\usepackage{booktabs}     
\usepackage{multirow}     
\usepackage{pifont}       
\usepackage{float}        
\usepackage{amsmath}      
\usepackage{amssymb}
\usepackage{hyperref}

\newcommand{\etal}{\textit{et~al.}}
\usepackage{orcidlink}
\hypersetup{
     colorlinks   = true,
     citecolor    = blue,
     linkcolor    = blue,
     urlcolor     = blue
}

\begin{document}
\title{ATCNet-CIAM for Multi-Session Motor Imagery EEG Signal Classification}
%
%
\author{
Le Huu Son Hai\inst {1} \and
Nguyen Chi Hai\inst {1} \and
Nguyen Thai Anh\inst {1}\orcidlink{0009-0005-5600-6510} \and
Tran Thien Thanh\inst {2}\orcidlink{0000-0002-1692-289X} \and
Vo Nguyen Quoc Bao\inst {1}\orcidlink{0000-0002-1791-6467} \and
Ngo Hoang Tu\inst {1}\thanks{Corresponding author: Ngo Hoang Tu (tu.nh@vlu.edu.vn).}\orcidlink{0000-0003-1944-6056}
}
\author{
Le Huu Son Hai\orcidlink{0009-0008-0293-4591} \and
Nguyen Chi Hai\orcidlink{0009-0001-0289-308X} \and
Truong Viet Vu\orcidlink{0009-0004-3941-1306} \and
Nguyen Phuc Nguyen\orcidlink{0009-0002-7337-337X} \and
Nguyen Thai Anh\orcidlink{0009-0005-5600-6510} \and
Ngo Hoang Tu\thanks{Corresponding author: Ngo Hoang Tu (tu.nh@vlu.edu.vn).}\orcidlink{0000-0003-1944-6056}
}
\authorrunning{L. H. S. Hai et al.}


%
\institute{\textsuperscript{1}Faculty of Information Technology, Van Lang School of Technology, Van Lang University, Ho Chi Minh City 70000, Vietnam \\
\textsuperscript{2}Department of Computer Engineering, Ho Chi Minh City University of Transport, Ho Chi Minh City 710372, Vietnam\\
\email{
hai.2274802010212@vanlanguni.vn,
hai.2274802010214@vanlanguni.vn,
anh.nt@vlu.edu.vn,
thanh.tran@ut.edu.vn,
bao.vnq@vlu.edu.vn,
tu.nh@vlu.edu.vn}
}
\institute{Faculty of Information Technology, Van Lang School of Technology, Van Lang University, Ho Chi Minh City 70000, Vietnam \\
\email{
hai.2274802010212@vanlanguni.vn,
hai.2274802010214@vanlanguni.vn,
vu.2274802011045@vanlanguni.vn,
nguyen.2274802010586@vanlanguni.vn,
anh.nt@vlu.edu.vn,
tu.nh@vlu.edu.vn}
}
\maketitle              

\begin{abstract}
Motor imagery (MI)-based electroencephalography is widely used in
non-invasive brain--computer interfaces (BCIs), but robust decoding
remains challenging due to inter-subject variability and cross-session
non-stationarity. This work proposes ATCNet-CIAM, an enhanced attention
temporal convolutional network that integrates a lightweight
channel-integrated attention module (CIAM) into the ATCNet framework
to improve channel-spatial feature representation for MI decoding.
The proposed model is evaluated on BCI Competition IV-2a, BCI
Competition IV-2b, and the multi-day WBCIC-MI dataset under standard,
within-session, and cross-session protocols. Experimental results show
that ATCNet-CIAM achieves 86.32\% accuracy on BCI IV-2a and 87.96\%
on BCI IV-2b under the standard protocol, while reaching 89.46\% and 83.64\% in
the within-session WBCIC-MI on 2C and 3C, respectively. The proposed framework
consistently improves classification stability and robustness under
session-varying conditions, and ablation study confirms the
complementary contribution of the proposed architectural components.
 
\end{abstract}
%
%

\section{Introduction}
 
Motor imagery (MI)-based electroencephalography (EEG) is a widely used
paradigm in non-invasive brain-computer interfaces (BCIs), enabling
users to interact with external systems through imagined motor activity
without physical movement. Despite substantial progress, reliable MI
decoding remains difficult due to low signal-to-noise ratio,
inter-subject variability, and cross-session EEG non-stationarity
\cite{tangermann2012review}. 
Recent surveys further
highlight that robust generalization across subjects and recording
sessions is still an open challenge for deep learning-based MI-BCI
systems \cite{hameed2025enhancing,saibene2024deep,al2021deep}.

Deep learning has gradually replaced handcrafted feature-engineering
pipelines for MI-EEG classification because of its ability to learn
discriminative spatio-temporal representations directly from raw EEG
signals \cite{lecun2015deep,roy2019deep}. Architectures such as
DeepConvNet \cite{schirrmeister2017deep}, EEGNet
\cite{lawhern2018eegnet}, and attention temporal convolutional network (ATCNet)
\cite{altaheri2022physics} demonstrated the effectiveness of temporal
convolution, spatial filtering, and attention-based modeling for MI
decoding. More recent studies have explored transformer-based and
multi-scale attention architectures to improve feature selectivity and
long-range temporal dependency modeling
\cite{MSVTNet}. Nevertheless, performance often
degrades under session-varying conditions, indicating that existing
models still struggle to learn robust EEG representations.

To address this issue, we propose ATCNet-CIAM, an enhanced attention
temporal convolutional network that integrates a lightweight
channel-integrated attention module (CIAM) into the ATCNet framework.
Unlike the original ATCNet, which applies attention primarily over sliding temporal windows, ATCNet-CIAM further introduces a frequency band attention (FBA) mechanism to adaptively emphasize informative EEG sub-band representations.
Inspired by the physiological relevance of rhythm-specific neural
activities in MI decoding \cite{pfurtscheller1999event}, FBA partitions intermediate feature channels into
multiple frequency-aware groups and dynamically estimates their relative
importance through learnable attention weights. This design enables the
network to selectively enhance discriminative spectral components while
suppressing less informative responses under inter-session variability.
ATCNet-CIAM introduces explicit channel-spatial feature refinement \emph{before} temporal aggregation
through four complementary components: (1) multi-scale temporal
convolution with parallel kernels of different receptive fields,
(2) a cross-gated channel-temporal interaction mechanism within
FBA-CIAM, and (3) efficient channel attention (ECA) integrated into dilated
temporal convolutional blocks and (4) adaptive window fusion (AWF) using attention-based aggregation across
sliding temporal windows. This design distinguishes the proposed
method from a simple extension of ATCNet and enables more discriminative
representation learning under session-varying conditions.
 
The proposed approach is evaluated on BCI Competition IV-2a
\cite{tangermann2012review} and the multi-day WBCIC-MI dataset
\cite{Yang2025} under standard, within-session, and cross-session
protocols. The focus of this work is robust MI decoding under standard
benchmark and session-varying conditions rather than subject-invariant
representation learning; cross-subject adaptation is therefore left as
future work. Experimental results show that ATCNet-CIAM consistently
improves MI classification performance and provides more stable results
under session-varying conditions.
The main contributions of this work are summarized as follows:
\begin{itemize}
    \item \textbf{We first propose ATCNet-CIAM}, a novel cross-gated channel-temporal attention module (FBA-CIAM) integrated into the ATCNet framework. The module couples frequency-band-aware channel refinement with multi-head temporal attention through a sigmoid gating mechanism, providing explicit channel-temporal interaction that goes beyond the window-level attention of the original ATCNet.
 
    \item \textbf{An enhanced ATCNet-CIAM architecture} is developed by combining multi-scale temporal convolution, FBA-CIAM, ECA-augmented dilated TCN, and adaptive window fusion within a unified lightweight framework, jointly addressing temporal dynamics, spatial selectivity, and adaptive channel weighting for MI-EEG decoding

 
    \item \textbf{Extensive experiments} on BCI Competition IV-2a, IV-2b, and WBCIC-MI demonstrate consistent improvements over strong baselines, with ATCNet-CIAM achieving 86.32\% on IV-2a, 87.96\% on IV-2b, and 89.46\%/83.64\% on WBCIC-MI 2C/3C within-session settings. Ablation studies further confirm the complementary contribution of each proposed component.
\end{itemize}

\section{Related Work}
\label{Sect:RelatedWork}
 
\textbf{Motor Imagery Brain-Computer Interfaces}:
MI-based BCIs have been widely 
studied for applications in neurorehabilitation, assistive systems, and 
human-computer interaction 
\cite{hameed2025enhancing,saibene2024deep,al2021deep}. 
MI-BCI systems decode imagined motor movements from EEG signals through 
modulations of sensorimotor rhythms, particularly event-related 
desynchronization and synchronization \cite{tarara2025motor}.
Recent surveys have summarized rapid progress in this field:
Hameed~\etal~\cite{hameed2025enhancing} reviewed convolutional neural networks (CNNs), recurrent networks,
temporal convolutional networks (TCNs), and transformer-based models; Saibene \etal~\cite{saibene2024deep}
compared subject-dependent and subject-independent evaluation protocols;
and Al-Saegh~\etal~\cite{al2021deep} highlighted the advantages of
deep feature learning over handcrafted methods. Despite these advances,
cross-session non-stationarity and inter-subject variability remain
open challenges \cite{tarara2025motor}.
 
\textbf{Deep Learning and Attention Mechanisms for MI-EEG}:
Deep learning has become a dominant approach for MI classification because of 
its ability to automatically learn hierarchical EEG representations 
\cite{lecun2015deep,roy2019deep}. 
Schirrmeister~\etal~\cite{schirrmeister2017deep} introduced DeepConvNet and 
ShallowConvNet, demonstrating the effectiveness of CNNs for end-to-end EEG 
decoding. EEGNet~\cite{lawhern2018eegnet} later proposed a lightweight 
architecture based on depthwise and separable convolutions, becoming one of 
the most widely used baselines in MI-BCI research.
Recent studies have further improved EEGNet and related architectures through 
attention mechanisms and hybrid CNN-transformer designs. 
AMEEGNet~\cite{wu2025ameegnet} integrated ECA to enhance discriminative feature learning, while MSVTNet~\cite{MSVTNet} employed 
multi-scale transformer modules for improved temporal-spatial modeling. 
Attention-based approaches have shown strong potential for emphasizing 
informative EEG channels and temporal dynamics while suppressing irrelevant 
features \cite{woo2018cbam,hou2025lightweight}.
 
\textbf{Cross-Session MI-EEG Decoding}: 
Cross-session and multi-day EEG decoding remain challenging because EEG 
distributions vary significantly across recording sessions and subjects 
\cite{Yang2025}. On the WBCIC-MI dataset, Yang~\etal~\cite{Yang2025} reported 
that conventional models such as EEGNet and DeepConvNet experience noticeable 
performance degradation under cross-session settings. To address this issue, 
previous studies have explored domain adaptation, contrastive learning, and 
feature alignment strategies 
\cite{liu2024cross,lotey2022cross}. 
Although these methods improve robustness, they often introduce additional 
training complexity and computational overhead. The present work focuses on 
architectural improvement within a subject-dependent framework rather than 
explicit domain adaptation; cross-subject generalization is an orthogonal 
direction that remains important future work.
 
\textbf{Research Gap and Motivation}: 
Although existing methods have achieved strong MI-EEG decoding performance,
they often struggle to jointly model temporal dynamics, spatial feature
selectivity, and adaptive channel weighting within a unified lightweight
framework. In addition, multi-day datasets such as WBCIC-MI remain
insufficiently explored for systematic cross-session evaluation.
Motivated by these gaps, this work proposes ATCNet-CIAM, which integrates
explicit channel-spatial attention and multi-scale temporal modeling into
ATCNet to improve both decoding accuracy and session-varying robustness.
 
\section{Proposed Methodology}
\label{Sect:method}
 
\subsection{Datasets}
 
Three benchmark datasets are used.
Table~\ref{tab:datasets} summarizes their specifications.
The model input is a tensor $(B, 1, C, T)$, where $B$ is batch size,
$1$ is a dummy channel required by Conv2D, $C$ is the number of EEG
channels, and $T$ is the number of time samples. This tensor is
permuted to $(B, T, C, 1)$ before entering the Multi-Scale Conv2D
block.
 
\begin{table}[t]
\centering
\caption{Dataset specifications and evaluation protocols used in this study.}
\label{tab:datasets}
\renewcommand{\arraystretch}{1.05}
\resizebox{\linewidth}{!}{
\begin{tabular}{@{}lcccccccc@{}}
\toprule
\textbf{Dataset} & \textbf{Ch.} & \textbf{Fs} &
\textbf{Cls.} & \textbf{Subj.} & \textbf{Sess.} &
\textbf{Trials/Sess.} & \textbf{Samples} & \textbf{Protocols used} \\
\midrule
BCI IV-2a  & 22 & 250 & 4 & 9  & 2 & 288      & 1125 & Standard \\
BCI IV-2b  & 3  & 250 & 2 & 9  & 5 & 120--160 & 1125 & Standard \\
WBCIC 2C   & 58 & 250 & 2 & 51 & 3 & 200      & 1000 & Standard, Within-sess., Cross-sess. \\
WBCIC 3C   & 58 & 250 & 3 & 11 & 3 & 300      & 1000 & Standard, Within-sess., Cross-sess. \\
\bottomrule
\end{tabular}
}
\end{table}
 
\textbf{BCI IV-2a}~\cite{tangermann2012review}: Nine subjects, four-class
MI (left hand, right hand, foot, tongue), 22 channels at 250\,Hz. The
analysis window spans 4.5\,s after cue onset ($t_1{=}1.5$\,s,
$t_2{=}6.0$\,s, $T{=}1125$ samples). Standard split: session~1 for
training, session~2 for testing.
 
\textbf{BCI IV-2b}~\cite{brunner2008bci}: Nine subjects, two-class MI
(left/right hand), three channels (C3, Cz, C4) at 250\,Hz; same
1125-sample window. Sessions~1--2 used for training,
sessions~3--5 for testing (standard competition protocol).
 
\textbf{WBCIC-MI}~\cite{Yang2025}: Multi-day Brain Imaging Data Structure (BIDS)-compliant benchmark;
51 subjects (2-class -- 2C) and 11 subjects (3-class -- 3C), 58 channels at 250\,Hz, 1000
samples per trial (4\,s). Three protocols are applied:
\emph{standard} (sessions 1+2 train / session~3 test),
\emph{within-session} (stratified 80/20 split within session~1), and
\emph{cross-session} (session~1 train / session~2 test, same subject,
different day).
 
\subsection{Data Preprocessing}
\label{ssec:preprocess}
 
All preprocessing follows a strict train-only normalization policy to
prevent data leakage.
 
\textbf{Epoch Extraction}:
For BCI IV-2a, each trial is extracted from the continuous EEG signal
using the onset sample provided in the \texttt{.mat} annotations. A
1750-sample raw window is first sliced from the trial onset, then
cropped to the MI-relevant interval
$[t_1{=}1.5\,\text{s},\, t_2{=}6.0\,\text{s}]$, resulting in
1125 samples per trial~\cite{tangermann2012review}. For BCI IV-2b,
the same 1125-sample window is obtained by cropping
$[t_1{=}3.0\,\text{s},\, t_2{=}7.5\,\text{s}]$ following the
competition protocol~\cite{brunner2008bci}. For WBCIC-MI, preprocessed
4\,s epochs ($T=1000$ samples at 250\,Hz) are loaded directly from the
BIDS-compliant derivatives archive~\cite{Yang2025}.
 
\textbf{Normalization and Reshape}:
Each trial is reshaped from $(C,T)$ to $(1,C,T)$ and stacked into a
four-dimensional tensor $(N,1,C,T)$ suitable for Conv2D input.
Per-channel z-score normalization is applied using statistics computed
exclusively from the training data and then applied to the corresponding
validation and test sets to prevent information leakage.
 
\textbf{Data Augmentation (Training Only)}:
Two augmentation strategies are applied during training only.
Sliding Window Augmentation randomly shifts each trial by up to
$\pm 50$ samples along the temporal axis while preserving the original
window length~\cite{lawhern2018eegnet}. In addition, within-class EEG
Mixup \cite{zhang2017mixup} linearly interpolates pairs of same-class
trials using mixing coefficients sampled from
$\text{Beta}(0.2,0.2)$. These augmentations improve training diversity
and robustness without modifying the evaluation protocol.
 
\subsection{ATCNet-CIAM Architecture}
\label{ssec:arch}
 
\begin{figure*}[t]
  \centering
  \includegraphics[width=\textwidth]{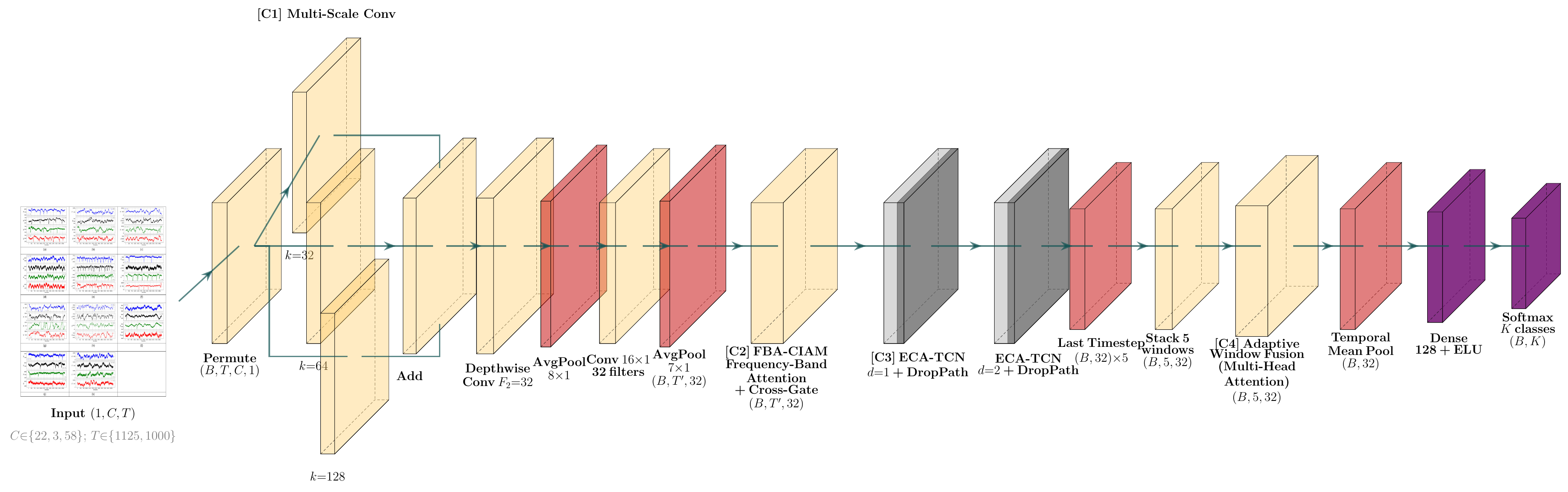}
  \caption{Overview of the proposed ATCNet-CIAM architecture. Input
  EEG tensor is processed by multi-scale temporal convolution, followed
  by parallel FBA-CIAM + ECA-TCN branches and adaptive window
  fusion for final classification.}
  \label{fig:architecture}
\end{figure*}

Fig.~\ref{fig:architecture} illustrates the overall ATCNet-CIAM
framework, whereas Table~\ref{tab:model_params} summarizes the main architectural
components of ATCNet-CIAM.
Details are discussed below.
 
\begin{table}[t]
\centering
\caption{Main architectural components of ATCNet-CIAM.}
\label{tab:model_params}
\renewcommand{\arraystretch}{1.05}
\scriptsize
\setlength{\tabcolsep}{3pt}
\begin{tabular}{@{}lp{4.5cm}p{4.1cm}@{}}
\toprule
\textbf{Block} &
\textbf{Main Components} &
\textbf{Purpose} \\
\midrule
Multi-Scale Conv
& Parallel temporal convolutions
& Multi-scale temporal feature extraction \\
 
FBA-CIAM
& Channel attention + multi-head attention + cross-gate
& Adaptive temporal-spatial refinement \\
 
ECA-TCN
& Dilated Conv1D + ECA + DropPath
& Temporal dependency modeling \\
 
AWF
& Attention-based window aggregation
& Window importance learning \\

\hline
Classification
& Dense + Softmax
& Final MI prediction \\
\bottomrule
\end{tabular}
\end{table}

\textbf{Multi-Scale Temporal Convolution:}
The input EEG tensor of shape $(B,T,C,1)$ is processed by three parallel temporal
Conv2D branches with kernel sizes $(32,1)$, $(64,1)$, and $(128,1)$,
each using same padding, unit stride, and $F_1 = 16$ filters, producing identical-shape feature maps $(B,T,C,F_1)$.
These correspond to receptive fields of
128\,ms, 256\,ms, and 512\,ms at 250\,Hz, providing implicit sensitivity to different temporal frequency bands.
The three maps are fused by element-wise addition, then passed through a depthwise convolution with kernel $(1, C)$ and depth multiplier $D=2$ to yield
$(B,T,1,F_2)$ with $F_2 = D \cdot F_1 = 32$.
A temporal average pooling $(8,1)$, a separable Conv2D $(16,1)$ with 32 filters, and a second pooling $(7,1)$ follow, producing the final tensor of shape $(B,T',F_2)$ with $T' = \lfloor T / 56 \rfloor$.
 
\textbf{FBA-CIAM:}
For each sliding window, the proposed FBA-CIAM module performs
multi-scale-aware channel refinement and temporal attention modeling.
Feature channels are divided into multiple groups corresponding to the
different temporal receptive fields of the preceding multi-scale
convolution and adaptively re-weighted using lightweight channel
attention. It is worth noting that this frequency-aware interpretation
is grounded in the implicit spectral sensitivity of multi-scale
temporal convolutions rather than explicit Fourier-domain
decomposition.
The refined features are then processed by pre-normalized multi-head attention with
residual connections. Finally, a cross-gate mechanism modulates the
temporal attention output using contextual channel information:
\begin{equation}
\mathbf{F}_{\mathrm{CIAM}}
=
\mathbf{F}_{\mathrm{ta}}
\odot
\sigma\big(
\mathbf{W}_g\,\mathbf{F}'_{[-1,:]}
\big),
\label{eq:crossgate}
\end{equation}
where $\mathbf{F}_{\mathrm{ta}}$ denotes the temporal attention
features, $\mathbf{F}'_{[-1,:]}$ is the channel-refined feature vector at the last temporal step,
$\mathbf{W}_g$ is a learnable projection matrix,
$\sigma(\cdot)$ is the sigmoid function,
and
$\odot$ denotes the Hadamard (element-wise) product, with the gating vector broadcast along the temporal dimension.
This cross-gate interaction
between channel and temporal pathways constitutes the key architectural
difference from the original ATCNet, which applies attention solely
over temporal windows without explicit channel-temporal coupling.
 
\textbf{ECA-TCN with DropPath:}
The attention-refined features are further processed using dilated
causal temporal convolutional blocks integrated with ECA~\cite{wang2020eca}. Stochastic Depth (DropPath)
regularization~\cite{huang2016deep} is additionally employed to
improve robustness and reduce overfitting on limited EEG data.
 
\textbf{AWF:}
The output features from all sliding windows are aggregated through
multi-head attention-based fusion to adaptively learn the relative
importance of each temporal window before final classification.

The novelty of this work lies in introducing an explicit cross-gated coupling between frequency-band-aware channel attention and temporal attention, a design motivated by the rhythm-specific nature of MI neural activity and shown to improve robustness under session-varying conditions. Architecturally, while ATCNet-CIAM retains ATCNet's convolutional encoder, sliding-window mechanism, and multi-head self-attention, it replaces the original single-scale temporal convolution, standard dilated TCN blocks, and concatenation-based window fusion with four enhanced components: multi-scale temporal convolution, frequency-band-aware channel attention with cross-gated channel-temporal fusion (i.e., FBA-CIAM), ECA-augmented dilated TCN, and AWF.
 
 
\subsection{Loss Function and Optimization}
 
The model is trained using categorical cross-entropy with label
smoothing ($\epsilon_s=0.05$)~\cite{szegedy2016rethinking}. Optimization
is performed using Adam with learning-rate warmup, gradient clipping,
and ReduceLROnPlateau scheduling. Early stopping is applied based on
validation accuracy to reduce overfitting. All experiments are
implemented in TensorFlow~2.x using mixed-precision training on an
NVIDIA RTX 4060 GPU.
\subsection{Evaluation Protocols}
 
For BCI IV-2a, session~1 is used for training and session~2 for testing \cite{tangermann2012review}. 
For BCI IV-2b, sessions~1--2 are used for training and sessions~3--5 for testing \cite{brunner2008bci}. For WBCIC-MI, three protocols are applied. The \emph{standard} protocol uses sessions~1--2 for training and session~3 for testing. The \emph{within-session} protocol applies a stratified 80/20 split within session~1, representing an idealized scenario with minimal temporal distribution shift. The \emph{cross-session} protocol trains on session~1 and tests on session~2 of the same subject, evaluating robustness under real recording-day variability. Standard evaluation is applied to all datasets; within-session and cross-session protocols are applied only to WBCIC-MI.
 
\textbf{Evaluation Metrics}:
Accuracy and Cohen's Kappa~\cite{cohen1960coefficient} are the primary metrics. Kappa is defined as $\kappa = (p_o - p_e)/(1 - p_e)$, where $p_o$ is observed agreement and $p_e$ is expected agreement by chance. Macro F1-score and confusion matrices are additionally reported for class-wise analysis.



\section{Experimental Results}
\label{sec:results}

\subsection{BCI Competition IV-2a}
 
Table~\ref{tab:comparison_iv2a} reports the standard-protocol results
on BCI IV-2a. ATCNet-CIAM achieved the highest mean accuracy (86.32\%) and the lowest standard deviation (6.50\%), outperforming
ATCNet~\cite{altaheri2022physics}, EEGNet~\cite{lawhern2018eegnet},
EEG-TCNet~\cite{ingolfsson2020eeg}, and
TCNet-Fusion~\cite{musallam2021electroencephalography}.
The improvement is particularly notable on Subject~2, where the
proposed model improves accuracy from 70.5\% to 80.9\%, suggesting
better robustness to noisy EEG patterns and subject variability.
Furthermore, Fig.~\ref{fig:all_confusion_matrices}(a) presents the aggregated confusion matrix,
showing balanced recognition across all four MI classes.
 
\begin{table}[t]
\centering
\caption{Per-subject accuracy (\%) and $\kappa$ on BCI IV-2a
(standard split).}
\label{tab:comparison_iv2a}
\renewcommand{\arraystretch}{1.05}
\scriptsize
\begin{tabular}{c|cc|cc|cc|cc|cc}
\toprule
\multirow{2}{*}{\textbf{Sub.}} &
\multicolumn{2}{c|}{\textbf{Ours}} &
\multicolumn{2}{c|}{\textbf{ATCNet~\cite{altaheri2022physics}}} &
\multicolumn{2}{c|}{\textbf{EEGNet~\cite{lawhern2018eegnet}}} &
\multicolumn{2}{c|}{\textbf{EEG-TCNet~\cite{ingolfsson2020eeg}}} &
\multicolumn{2}{c}{\textbf{TCNet-F~\cite{musallam2021electroencephalography}}} \\
& \% & $\kappa$ & \% & $\kappa$ & \% & $\kappa$ & \% & $\kappa$ & \% & $\kappa$ \\
\midrule
1 & 89.8 & 0.86 & 88.5 & 0.85 & 88.5 & 0.85 & 84.0 & 0.79 & 86.1 & 0.81 \\
2 & 80.9 & 0.75 & 70.5 & 0.61 & 66.0 & 0.55 & 66.3 & 0.55 & 66.0 & 0.55 \\
3 & 96.2 & 0.95 & 97.6 & 0.97 & 95.1 & 0.94 & 94.1 & 0.92 & 93.4 & 0.91 \\
4 & 79.7 & 0.73 & 81.0 & 0.75 & 73.6 & 0.65 & 72.6 & 0.63 & 72.6 & 0.63 \\
5 & 83.9 & 0.78 & 83.0 & 0.77 & 75.4 & 0.67 & 76.0 & 0.68 & 79.9 & 0.73 \\
6 & 75.0 & 0.67 & 73.6 & 0.65 & 64.2 & 0.52 & 62.9 & 0.50 & 66.7 & 0.56 \\
7 & 91.8 & 0.89 & 93.1 & 0.91 & 90.3 & 0.87 & 89.9 & 0.87 & 90.3 & 0.87 \\
8 & 91.0 & 0.88 & 90.3 & 0.87 & 85.8 & 0.81 & 84.7 & 0.80 & 85.8 & 0.81 \\
9 & 88.7 & 0.85 & 91.0 & 0.88 & 86.5 & 0.82 & 85.4 & 0.81 & 85.4 & 0.81 \\
\midrule
{Mean} &
\textbf{86.32} & \textbf{0.82} &
85.4 & 0.81 &
80.6 & 0.74 &
79.6 & 0.73 &
80.7 & 0.74 \\
 
{Std} &
\textbf{6.50} & \textbf{0.09} &
9.1 & 0.12 &
11.1 & 0.15 &
10.7 & 0.14 &
10.1 & 0.13 \\
\bottomrule
\end{tabular}
\end{table}

\subsection{BCI Competition IV-2b}
 
Table~\ref{tab:comparison_iv2b} summarizes the results on BCI IV-2b.
ATCNet-CIAM achieved the best mean accuracy (87.96\%) and competitive
$\kappa$ values, outperforming both classical and deep-learning
baselines including FBCSP~\cite{ang2008filter},
EEGNet~\cite{lawhern2018eegnet},
DRDA~\cite{zhao2020deep}, and EEG-DG~\cite{zhong2024eeg}.
The results are notable because IV-2b contains only three EEG channels,
making spatial discrimination substantially more difficult.
The aggregated confusion matrix in Fig.~\ref{fig:all_confusion_matrices}(b) further confirms balanced two-class recognition, with left-hand and right-hand classes achieving 0.90 and 0.86 recall respectively, indicating that the proposed model handles binary MI discrimination consistently even with only three EEG channels.
 
\begin{table}[t]
\centering
\caption{Per-subject accuracy (\%) and $\kappa$ on BCI IV-2b
(standard split).}
\label{tab:comparison_iv2b}
\renewcommand{\arraystretch}{1.1}
\resizebox{\linewidth}{!}{
\begin{tabular}{c|cc|cc|cc|cc|cc|cc|cc|cc}
\toprule
\multirow{2}{*}{\scriptsize \textbf{Sub.}} &
\multicolumn{2}{c|}{\textbf{Ours}} &
\multicolumn{2}{c|}{\scriptsize \textbf{FBCSP~\cite{ang2008filter}}} &
\multicolumn{2}{c|}{\scriptsize \textbf{CCSP~\cite{kang2009composite}}} &
\multicolumn{2}{c|}{\scriptsize \textbf{EEGNet~\cite{lawhern2018eegnet}}} &
\multicolumn{2}{c|}{\scriptsize \textbf{ConvNet~\cite{schirrmeister2017deep}}} &
\multicolumn{2}{c|}{\scriptsize \textbf{GAT~\cite{song2023global}}} &
\multicolumn{2}{c|}{\scriptsize \textbf{DRDA~\cite{zhao2020deep}}} &
\multicolumn{2}{c}{\scriptsize \textbf{EEG-DG~\cite{zhong2024eeg}}} \\
& \% & $\kappa$ & \% & $\kappa$ & \% & $\kappa$ & \% & $\kappa$ & \% & $\kappa$ & \% & $\kappa$ & \% & $\kappa$ & \% & $\kappa$ \\
\midrule
1 & 80.00 & 0.6000 & 70.00 & -- & 63.75 & -- & 70.31 & -- & 76.56 & -- & 84.58 & -- & 81.37 & -- & 82.50 & -- \\
2 & 70.36 & 0.4071 & 60.36 & -- & 56.79 & -- & 70.36 & -- & 50.00 & -- & 61.67 & -- & 62.86 & -- & 67.50 & -- \\
3 & 85.62 & 0.7125 & 60.94 & -- & 50.00 & -- & 78.44 & -- & 51.56 & -- & 60.83 & -- & 63.63 & -- & 72.19 & -- \\
4 & 98.44 & 0.9688 & 97.50 & -- & 93.44 & -- & 95.33 & -- & 96.88 & -- & 99.58 & -- & 95.94 & -- & 98.44 & -- \\
5 & 96.56 & 0.9313 & 93.12 & -- & 65.63 & -- & 93.44 & -- & 93.13 & -- & 87.50 & -- & 93.56 & -- & 96.56 & -- \\
6 & 87.19 & 0.7438 & 80.63 & -- & 81.25 & -- & 82.18 & -- & 85.31 & -- & 93.33 & -- & 88.19 & -- & 90.94 & -- \\
7 & 91.25 & 0.8250 & 78.13 & -- & 72.81 & -- & 91.88 & -- & 83.75 & -- & 85.42 & -- & 85.00 & -- & 89.38 & -- \\
8 & 93.75 & 0.8750 & 92.50 & -- & 87.81 & -- & 87.19 & -- & 91.56 & -- & 95.00 & -- & 95.25 & -- & 95.00 & -- \\
9 & 88.44 & 0.7688 & 86.88 & -- & 82.81 & -- & 71.65 & -- & 85.62 & -- & 92.08 & -- & 90.00 & -- & 91.56 & -- \\
\midrule
\textbf{Mean} & \textbf{87.96} & \textbf{0.7591} & 80.00 & -- & 72.70 & -- & 82.37 & -- & 79.37 & -- & 84.44 & -- & 83.98 & -- & 87.12 & -- \\
\textbf{St.D.} & \textbf{8.22} & \textbf{0.1643} & 13.85 & -- & 14.72 & -- & 10.15 & -- & 17.26 & -- & 13.98 & -- & 12.67 & -- & 10.89 & -- \\
\bottomrule
\end{tabular}
}
\end{table}

\subsection{WBCIC-MI Robustness Evaluation}
 
\begin{figure*}[t]
  \centering

  \begin{subfigure}[t]{0.3\textwidth}
    \centering
    \includegraphics[width=\linewidth]{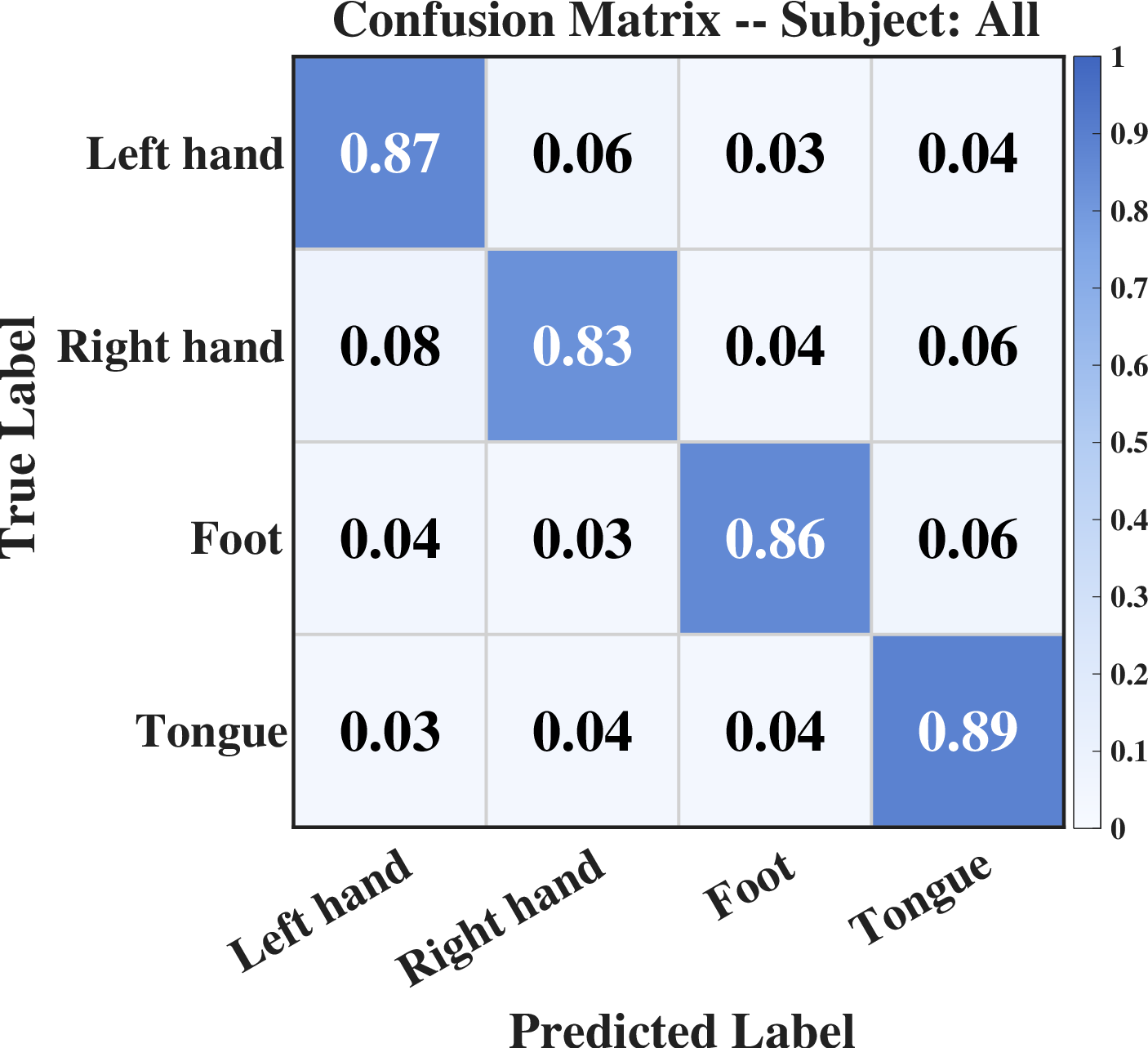}
    \caption{BCI IV-2a}
  \end{subfigure}
  \begin{subfigure}[t]{0.3\textwidth}
    \centering
    \includegraphics[width=\linewidth]{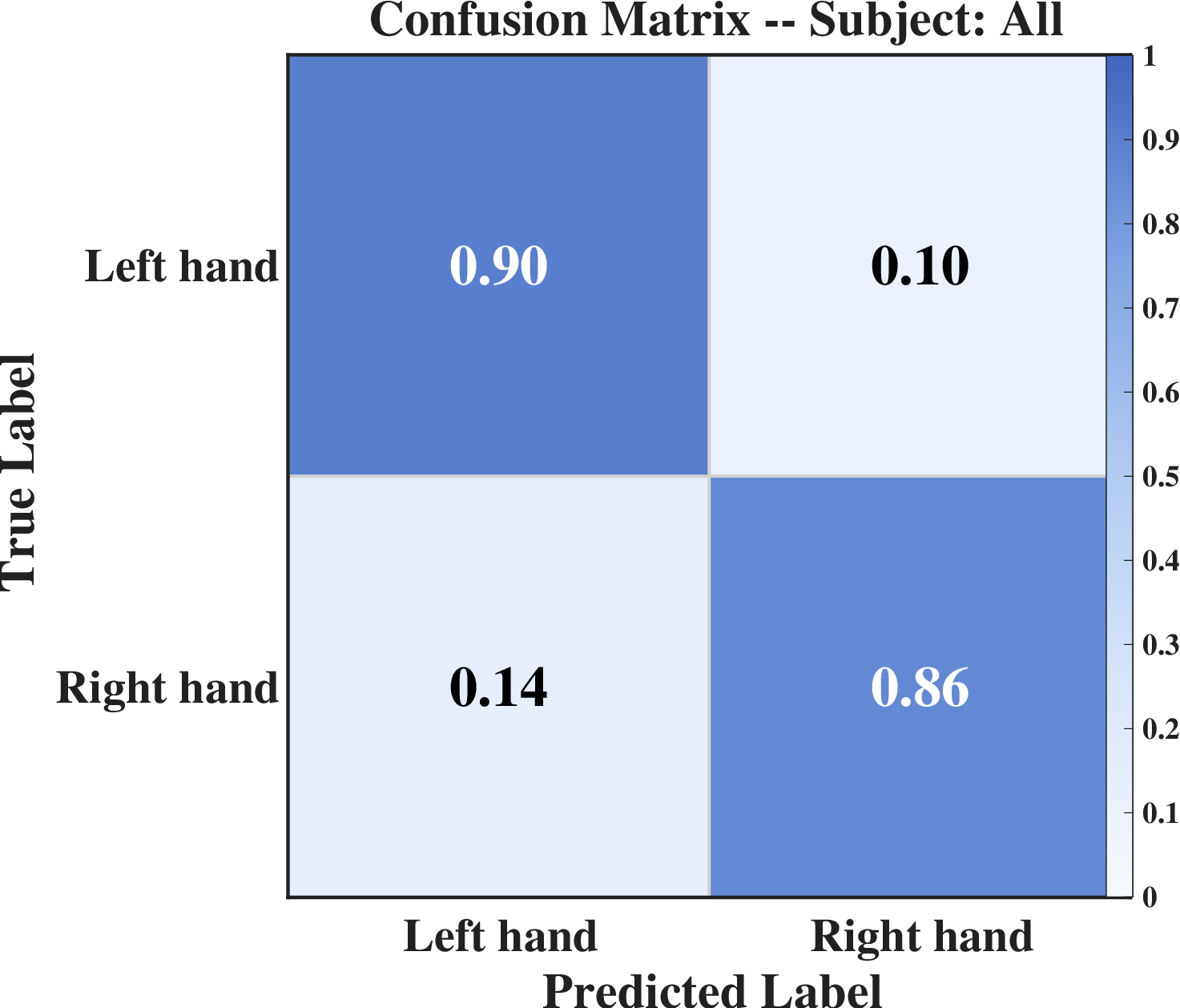}
    \caption{BCI IV-2b}
  \end{subfigure}

  \vspace{3pt}

  \begin{subfigure}[t]{0.3\textwidth}
    \centering
    \includegraphics[width=\linewidth]{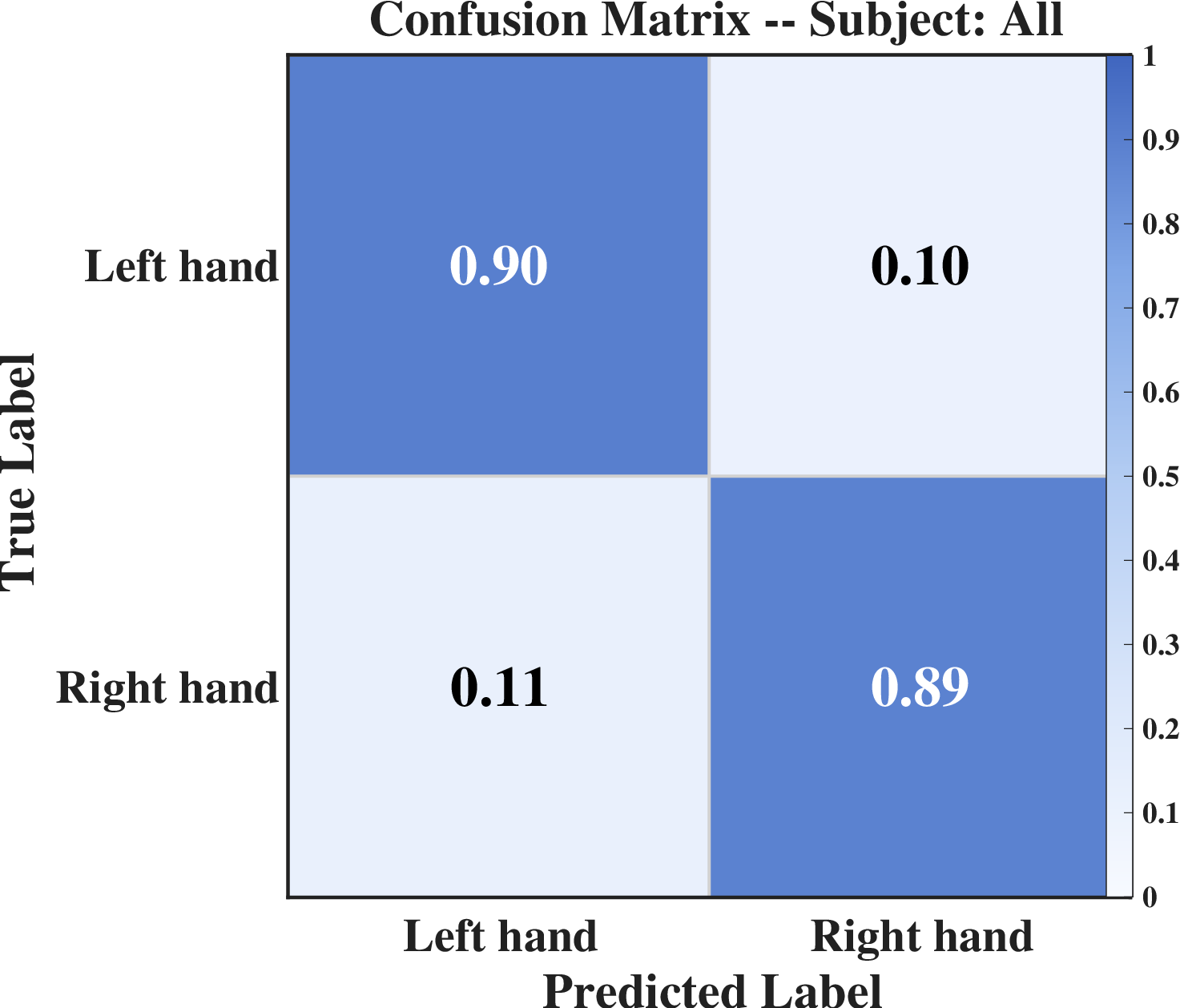}
    \caption{2C within-session}
  \end{subfigure}
  \hfill
  \begin{subfigure}[t]{0.3\textwidth}
    \centering
    \includegraphics[width=\linewidth]{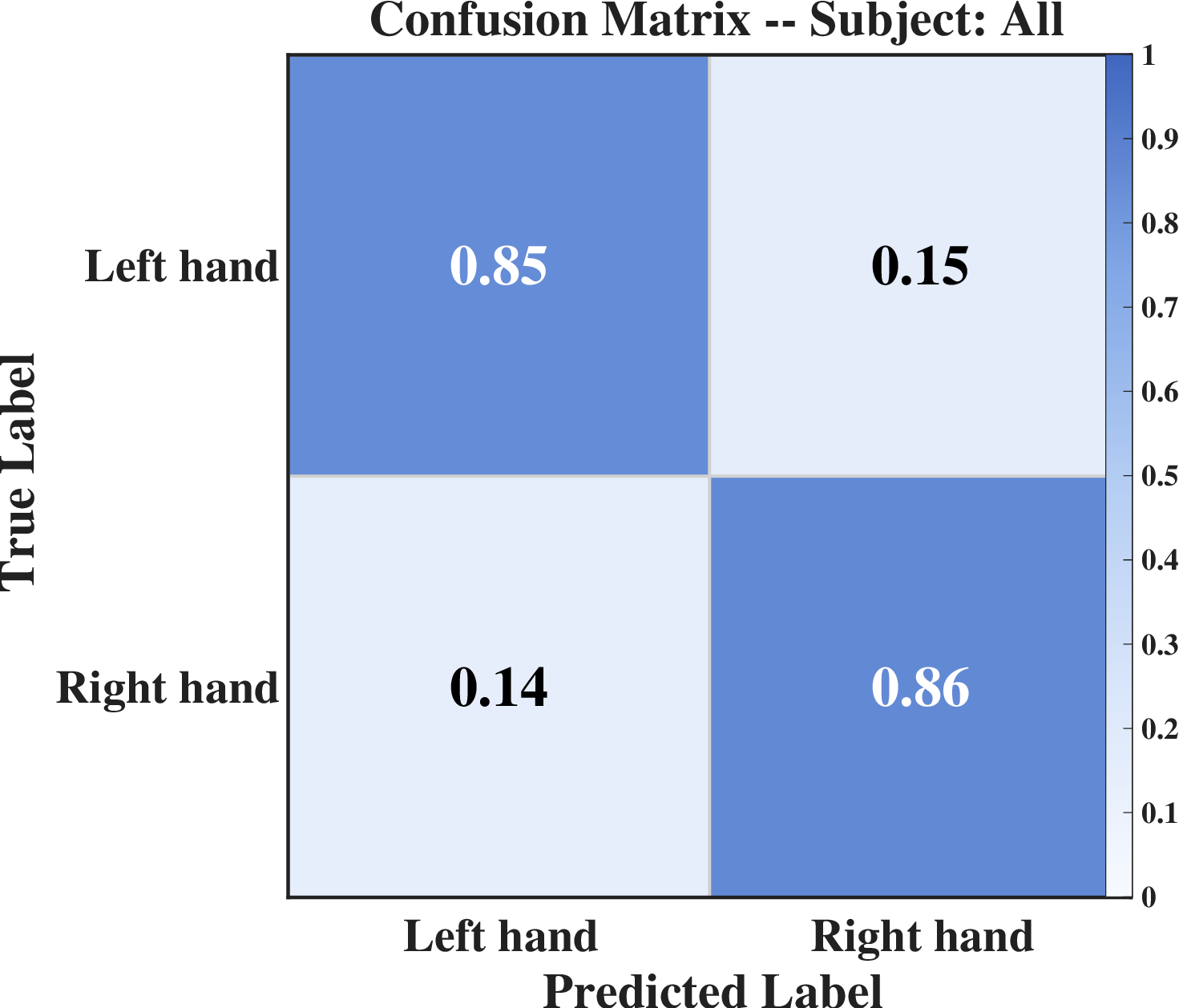}
    \caption{2C standard}
  \end{subfigure}
    \hfill
  \begin{subfigure}[t]{0.3\textwidth}
    \centering
    \includegraphics[width=\linewidth]{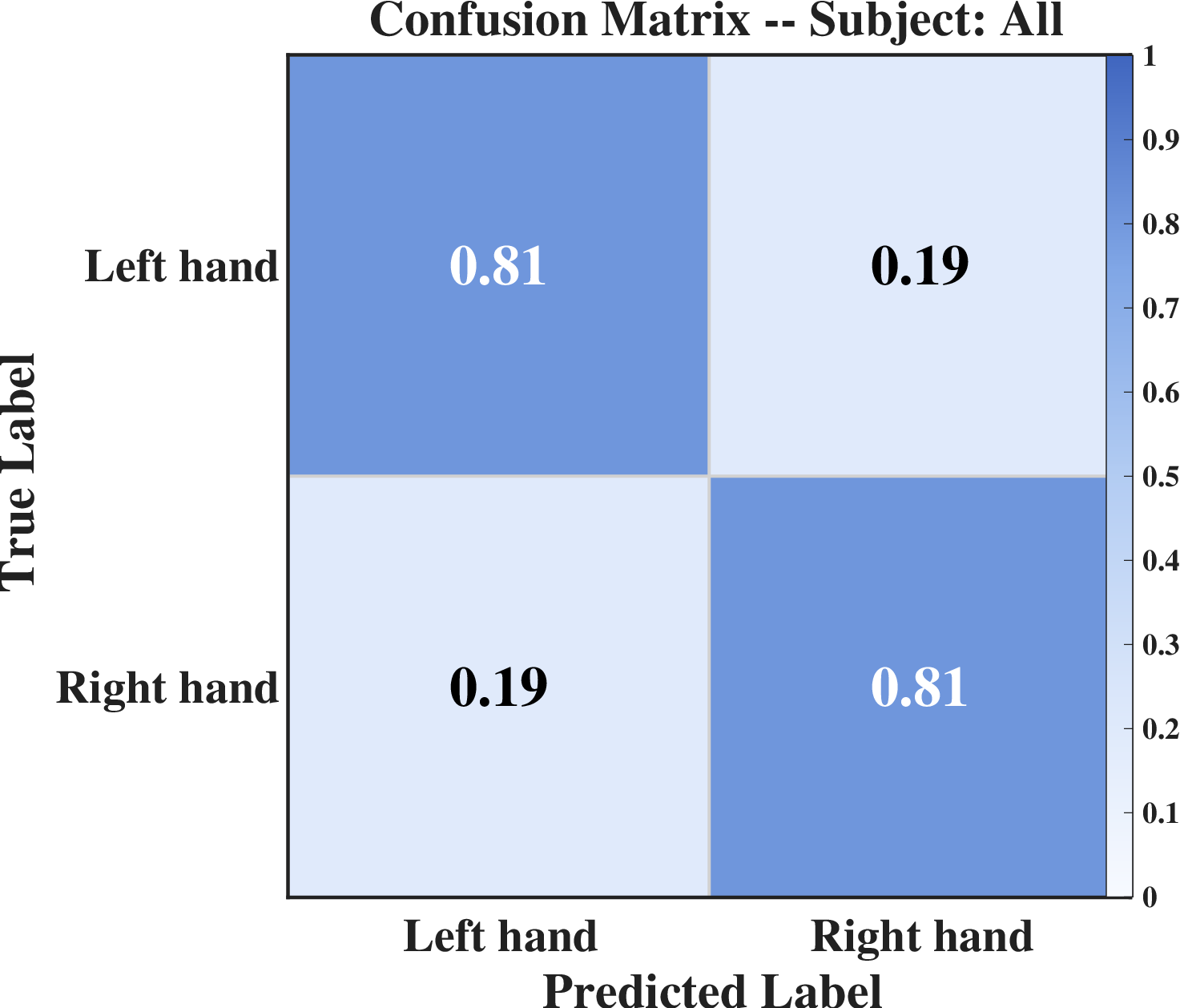}
    \caption{2C cross-session}
  \end{subfigure}


  \vspace{3pt}

  \begin{subfigure}[t]{0.3\textwidth}
    \centering
    \includegraphics[width=\linewidth]{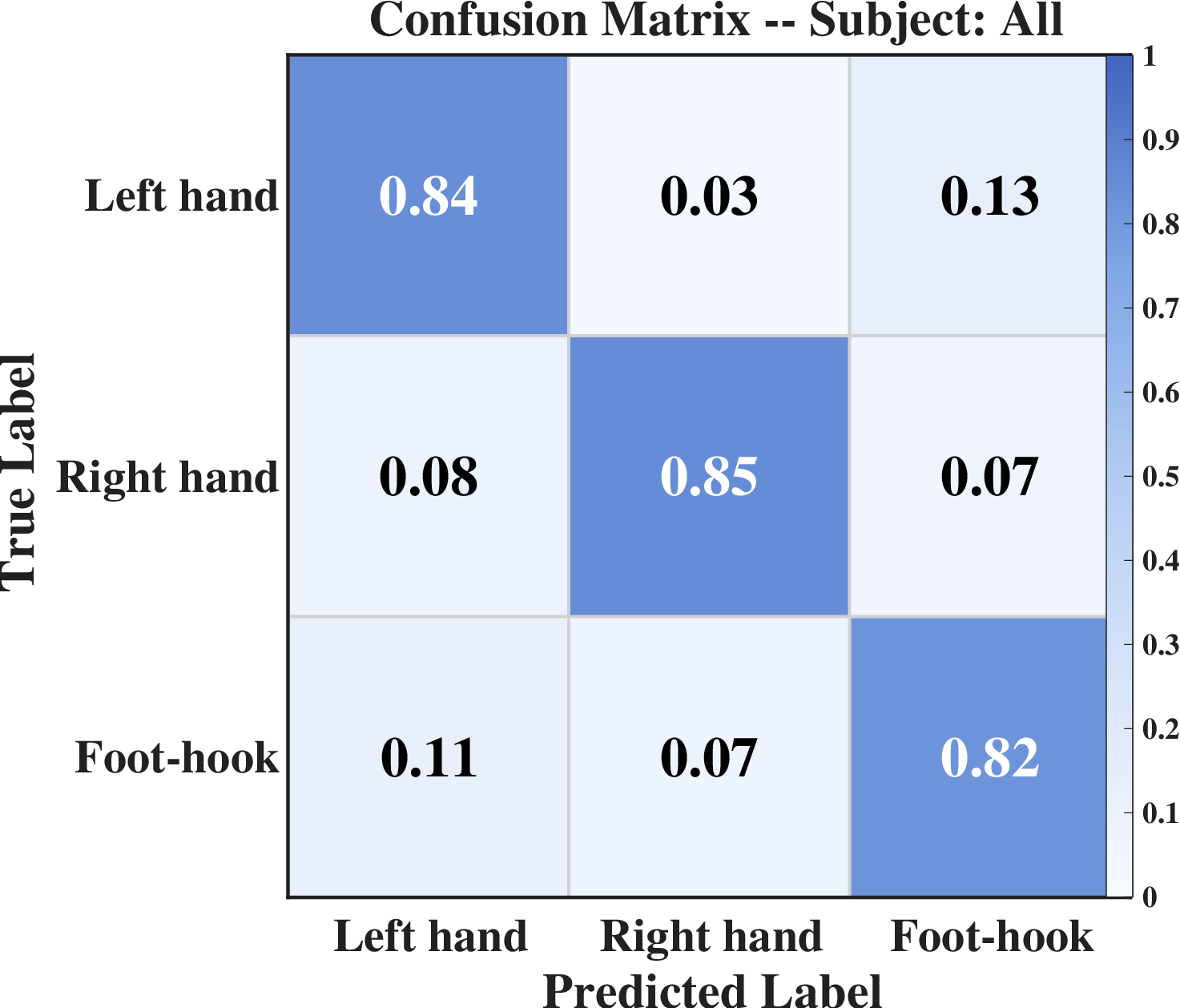}
    \caption{3C within-session}
  \end{subfigure}
   \hfill
  \begin{subfigure}[t]{0.3\textwidth}
    \centering
    \includegraphics[width=\linewidth]{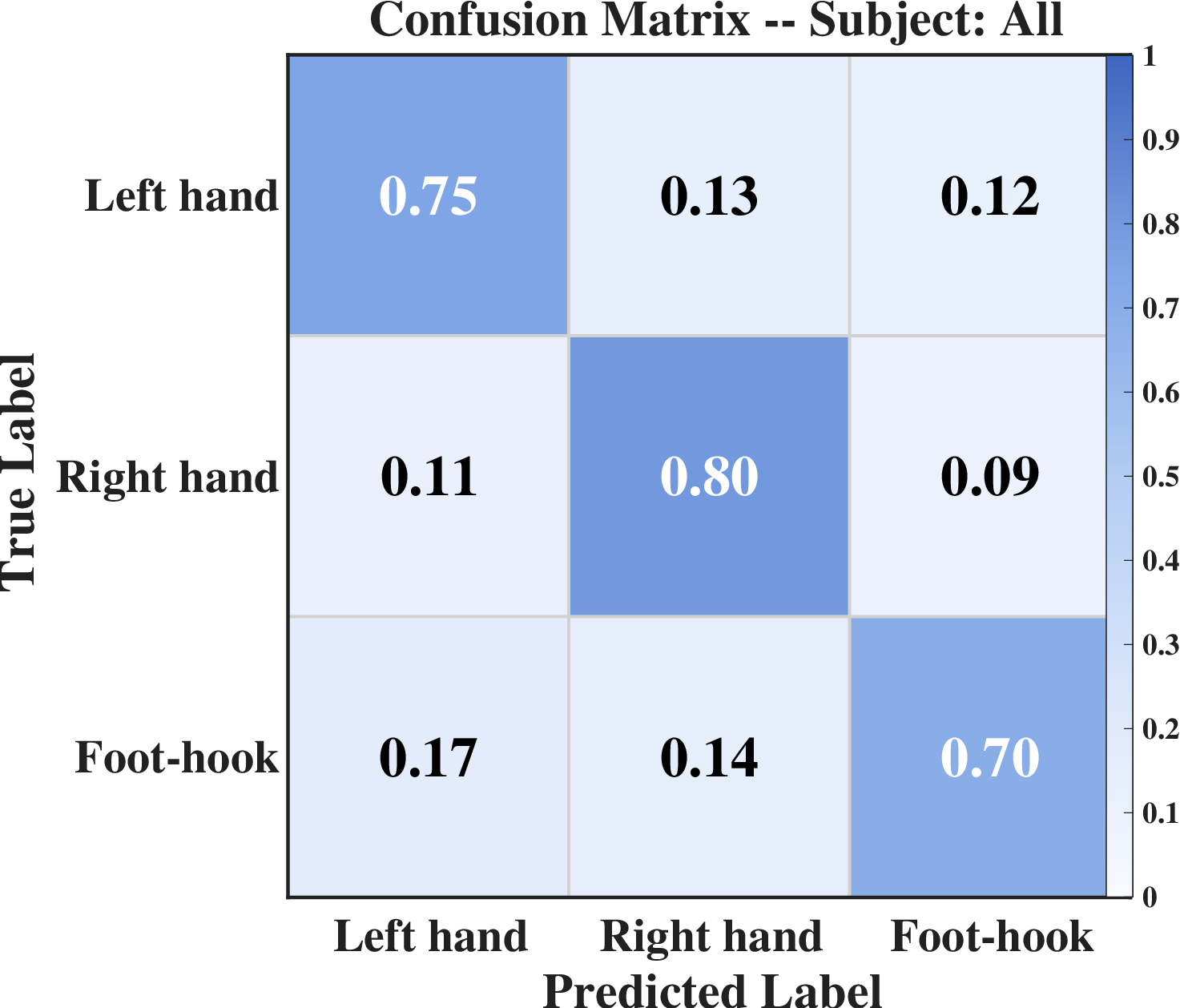}
    \caption{3C standard}
  \end{subfigure}
  \hfill
  \begin{subfigure}[t]{0.3\textwidth}
    \centering
    \includegraphics[width=\linewidth]{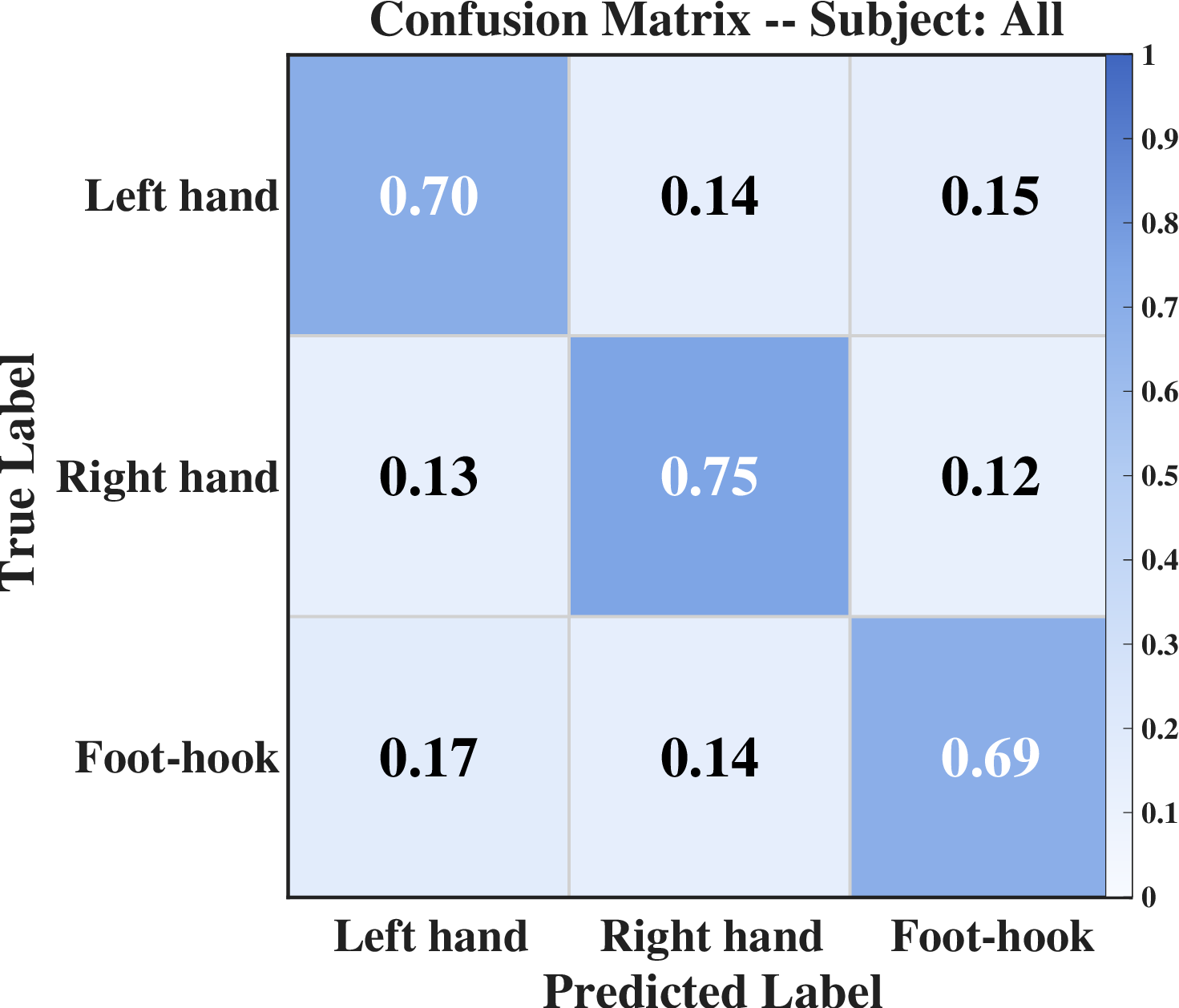}
    \caption{3C cross-session}
  \end{subfigure}
  \caption{Aggregated confusion matrices across BCI IV-2a, BCI IV-2b, and WBCIC-MI datasets under different evaluation protocols.}
  \label{fig:all_confusion_matrices}
\end{figure*}

Table~\ref{tab:wbcic_all} reports the results on the multi-day
WBCIC-MI benchmark under standard, within-session, and cross-session
protocols. Within-session evaluation achieved the highest performance
for both paradigms, reaching 89.46\% in 2C and 83.64\% in 3C,
indicating the upper performance bound when session-level distribution
shift is absent.
The corresponding confusion matrices in Fig.~\ref{fig:all_confusion_matrices}(c) and Fig.~\ref{fig:all_confusion_matrices}(f) show highly balanced per-class recognition, with all diagonal entries above 0.82.
Under the standard protocol, the confusion matrices in Fig.~\ref{fig:all_confusion_matrices}(d) (2C) and Fig.~\ref{fig:all_confusion_matrices}(g) (3C) remain well-balanced, although the 3C setting exhibits a noticeable drop on the foot-hook class (recall 0.70), reflecting the inherent difficulty of three-class MI discrimination.

Performance decreases under cross-session evaluation, as illustrated by Fig.~\ref{fig:all_confusion_matrices}(e) for 2C and Fig.~\ref{fig:all_confusion_matrices}(h) for 3C, where the diagonal entries drop to 0.81 for both 2C classes and to 0.69--0.75 across the 3C classes. This degradation reflects the difficulty introduced by recording-day variability and EEG non-stationarity. Nevertheless, ATCNet-CIAM maintains relatively stable performance across all settings, suggesting that the proposed frequency-aware attention and multi-scale temporal modeling improve robustness under session-varying conditions.

 
\begin{table}[t]
\centering
\caption{ATCNet-CIAM performance on WBCIC-MI (mean $\pm$ std).}
\label{tab:wbcic_all}
\renewcommand{\arraystretch}{1.0}
\scriptsize
\begin{tabular}{@{}llcc@{}}
\toprule
\textbf{Paradigm} & \textbf{Protocol} &
\textbf{Accuracy (\%)} & \textbf{$\kappa$} \\
\midrule
\multirow{3}{*}{2C (51 subjects)}
  & Standard       & 85.39 $\pm$ 12.93 & 0.708 $\pm$ 0.259 \\
  & Within-session & \textbf{89.46} $\pm$ 10.47 & \textbf{0.789} $\pm$ 0.209 \\
  & Cross-session  & 81.28 $\pm$ 11.81 & 0.626 $\pm$ 0.236 \\
\midrule
\multirow{3}{*}{3C (11 subjects)}
  & Standard       & 74.88 $\pm$ 13.38 & 0.623 $\pm$ 0.201 \\
  & Within-session & \textbf{83.64} $\pm$ 7.38 & \textbf{0.755} $\pm$ 0.111 \\
  & Cross-session  & 71.36 $\pm$ 12.99 & 0.571 $\pm$ 0.195 \\
\bottomrule
\end{tabular}
\end{table}
 
\subsection{Ablation Study on BCI IV-2a}
\label{sec:ablation}

\begin{table}[t]
\centering
\caption{Ablation study on BCI IV-2a (standard protocol).}
\label{tab:ablation}
\renewcommand{\arraystretch}{1.2}
\setlength{\tabcolsep}{4pt}
\scriptsize
\begin{tabular}{@{}lcc@{}}
\toprule
\textbf{Configuration} & \textbf{Accuracy} &\textbf{$\kappa$} \\
\midrule
ATCNet-CIAM (full model)    & \textbf{86.32}\% & \textbf{0.818} \\
Without AWF         & 76.50\% $(\downarrow 9.82\%)$ & 0.687 $(\downarrow 16.01\%)$ \\
Without FBA-CIAM                & 78.63\% $(\downarrow 7.69\%)$ & 0.715 $(\downarrow 12.59\%)$ \\
Without Multi-Scale Conv        & 79.48\% $(\downarrow 6.84\%)$ & 0.726 $(\downarrow 11.25\%)$ \\
Without ECA-TCN                 & 79.40\% $(\downarrow 6.92\%)$ & 0.725 $(\downarrow 11.37\%)$ \\
\bottomrule
\end{tabular}
\end{table}

Table~\ref{tab:ablation} presents the ablation results on BCI IV-2a under the standard evaluation protocol.
Removing any component leads to a noticeable performance drop,
indicating that the proposed modules contribute complementary
information. The largest degradation is observed when Adaptive Window
Fusion is removed, suggesting that attention-based aggregation is
important for preserving discriminative temporal context across
sliding windows. Removing FBA-CIAM also causes a substantial decrease
in performance, supporting the effectiveness of the proposed
channel-temporal cross-gate mechanism for MI-EEG representation learning.
In comparison with the original ATCNet baseline, the full ATCNet-CIAM
model improves both accuracy and kappa, demonstrating the benefit of
integrating multi-scale temporal modeling and lightweight attention
refinement beyond what ATCNet's window-level attention alone provides.
 
 
\section{Discussion}
\label{sec:discussion}
 
The experimental results demonstrate that ATCNet-CIAM consistently
improves MI-EEG decoding performance under both standard and
session-varying evaluation settings. On BCI IV-2a and IV-2b, the proposed model achieves higher accuracy and kappa scores than the baseline ATCNet, indicating that the integration of channel-spatial
attention and adaptive temporal fusion improves discriminative feature
learning. The key architectural novelty relative to ATCNet lies in
the cross-gate mechanism within FBA-CIAM, which couples channel-refined
features with the temporal attention pathway. Unlike ATCNet's
window-level attention, this coupling enables the model to jointly
refine channel selectivity and temporal context, resulting in more
stable per-subject performance as reflected by the lower standard
deviation on BCI IV-2a (6.5 vs.\ 9.1 for ATCNet).
 
The cross-session experiments on WBCIC-MI further show that session
variability remains a major challenge for practical MI-BCI systems.
The observed performance gap between within-session and cross-session
evaluation is an expected consequence of EEG non-stationarity across
recording days: electrode placement variability, impedance differences,
and changes in subject mental state introduce distribution shifts that
architectural improvements alone cannot fully compensate for.
Nevertheless, ATCNet-CIAM narrows this gap relative to the baseline,
suggesting that the proposed multi-scale temporal modeling and
lightweight channel attention improve feature stability under
session-varying conditions. Fully resolving cross-session non-stationarity
is more likely to require explicit domain adaptation or feature alignment
strategies, which are orthogonal to the present contribution and remain
important directions for future work.

Despite these improvements, several limitations remain. The framework relies on relatively long EEG analysis windows, restricting real-time deployment; the model is evaluated only under subject-dependent settings without addressing cross-subject generalization; substantial degradation persists under cross-session evaluation, indicating that EEG non-stationarity is not fully resolved; and computational latency and energy efficiency were not systematically assessed. Future work will therefore investigate lightweight adaptation, efficient inference, and domain generalization techniques for practical online BCI systems.

 
 
\section{Conclusion}
\label{sec:conclusion}
 
This study presented ATCNet-CIAM, an enhanced temporal-spatial
architecture for MI-EEG classification that integrates lightweight
channel-spatial attention and adaptive temporal fusion into the
ATCNet framework. The central innovation is the FBA-CIAM module,
which introduces a cross-gate mechanism that explicitly couples
channel-refined features with the temporal attention pathway,
a design that goes beyond the window-level attention of the original
ATCNet. Experimental results on BCI IV-2a, IV-2b, and the
multi-day WBCIC-MI benchmark demonstrate consistent improvements under
standard, within-session, and cross-session evaluation settings.
Ablation study confirms that the proposed modules contribute
complementary benefits for robust MI representation learning.
Overall, the results suggest that combining multi-scale temporal
modeling with lightweight channel-temporal attention refinement is an
effective direction for improving practical MI-BCI decoding performance.
Future work will focus on session adaptation, cross-subject
generalization, and computationally efficient deployment for
real-time applications.

%
%
\bibliographystyle{splncs04}
\bibliography{mybibliography}

@article{hameed2025enhancing,
  title={Enhancing motor imagery EEG signal decoding through machine learning: A systematic review of recent progress},
  author={Hameed, Ibtehaaj and Khan, Danish M and Ahmed, Syed Muneeb and Aftab, Syed Sabeeh and Fazal, Hammad},
  journal={Computers in Biology and Medicine},
  volume={185},
  pages={109534},
  year={2025},
  publisher={Elsevier}
}

@article{saibene2024deep,
  title={Deep learning in motor imagery EEG signal decoding: A Systematic Review},
  author={Saibene, Aurora and Ghaemi, Hafez and Dagdevir, Eda},
  journal={Neurocomputing},
  volume={610},
  pages={128577},
  year={2024},
  publisher={Elsevier}
}

@article{al2021deep,
  title={Deep learning for motor imagery EEG-based classification: A review},
  author={Al-Saegh, Ali and Dawwd, Shefa A and Abdul-Jabbar, Jassim M},
  journal={Biomedical Signal Processing and Control},
  volume={63},
  pages={102172},
  year={2021},
  publisher={Elsevier}
}

@inproceedings{huang2016deep,
  title={Deep networks with stochastic depth},
  author={Huang, Gao and Sun, Yu and Liu, Zhuang and Sedra, Daniel and Weinberger, Kilian Q},
  booktitle={European conference on computer vision},
  pages={646--661},
  year={2016},
  organization={Springer}
}

@inproceedings{wang2020eca,
  title={ECA-Net: Efficient channel attention for deep convolutional neural networks},
  author={Wang, Qilong and Wu, Banggu and Zhu, Pengfei and Li, Peihua and Zuo, Wangmeng and Hu, Qinghua},
  booktitle={Proceedings of the IEEE/CVF conference on computer vision and pattern recognition},
  pages={11534--11542},
  year={2020}
}

@inproceedings{szegedy2016rethinking,
  title={Rethinking the inception architecture for computer vision},
  author={Szegedy, Christian and Vanhoucke, Vincent and Ioffe, Sergey and Shlens, Jon and Wojna, Zbigniew},
  booktitle={Proceedings of the IEEE conference on computer vision and pattern recognition},
  pages={2818--2826},
  year={2016}
}

@article{zhang2017mixup,
  title={mixup: Beyond empirical risk minimization},
  author={Zhang, Hongyi and Cisse, Moustapha and Dauphin, Yann N and Lopez-Paz, David},
  journal={arXiv preprint arXiv:1710.09412},
  year={2017}
}

@article{tarara2025motor,
title={Motor imagery-based brain-computer interfaces: an exploration of multiclass motor imagery-based control for {E}motiv {EPOC} {X}},
author={Tarara, Paulina and Przyby{\l}, Iwona and Sch{\"o}ning, Julius and Gunia, Artur},
journal={Frontiers in Neuroinformatics},
volume={19},
pages={1625279},
year={2025},
publisher={Frontiers}
}

@article{Yang2025,
author = {Yang, Banghua and Rong, Fenqi and Xie, Yunlong and Li, Du and Zhang, Jiayang and Li, Fu and Shi, Guangming and Gao, Xiaorong},
year = {2025},
month = {03},
pages = {},
title = {A multi-day and high-quality {EEG} dataset for motor imagery brain-computer interface},
volume = {12},
journal = {Scientific Data},
doi = {10.1038/s41597-025-04826-y}
}

@article{schirrmeister2017deep,
title={Deep learning with convolutional neural networks for {EEG} decoding and visualization},
author={Schirrmeister, Robin Tibor and Springenberg, Jost Tobias and Fiederer, Lukas Dominique Josef and Glasstetter, Martin and Eggensperger, Katharina and Tangermann, Michael and Hutter, Frank and Burgard, Wolfram and Ball, Tonio},
journal={Human brain mapping},
volume={38},
number={11},
pages={5391--5420},
year={2017},
publisher={Wiley Online Library}
}

@article{lawhern2018eegnet,
title={{EEGNet}: a compact convolutional neural network for {EEG}-based brain--computer interfaces},
author={Lawhern, Vernon J and Solon, Amelia J and Waytowich, Nicholas R and Gordon, Stephen M and Hung, Chou P and Lance, Brent J},
journal={Journal of neural engineering},
volume={15},
number={5},
pages={056013},
year={2018},
publisher={iOP Publishing}
}

@article{wu2025ameegnet,
title={{AMEEGNet}: attention-based multiscale {EEGNet} for effective motor imagery {EEG} decoding},
author={Wu, Xuejian and Chu, Yaqi and Li, Qing and Luo, Yang and Zhao, Yiwen and Zhao, Xingang},
journal={Frontiers in Neurorobotics},
volume={19},
pages={1540033},
year={2025},
publisher={Frontiers Media SA}
}

@ARTICLE{MSVTNet,
author={Liu, Ke and Yang, Tao and Yu, Zhuliang and Yi, Weibo and Yu, Hong and Wang, Guoyin and Wu, Wei},
journal={IEEE Journal of Biomedical and Health Informatics},
title={{MSVTNet}: Multi-Scale Vision Transformer Neural Network for {EEG}-Based Motor Imagery Decoding},
year={2024},
volume={28},
number={12},
pages={7126-7137},
doi={10.1109/JBHI.2024.3450753}}

@article{hou2025lightweight,
  title={A lightweight convolutional transformer neural network for EEG-based depression recognition},
  author={Hou, Pengfei and Li, Xiaowei and Zhu, Jing and Hu, Bin},
  journal={Biomedical Signal Processing and Control},
  volume={100},
  pages={107112},
  year={2025},
  publisher={Elsevier}
}

@article{altaheri2022physics,
title={Physics-informed attention temporal convolutional network for {EEG}-based motor imagery classification},
author={Altaheri, Hamdi and Muhammad, Ghulam and Alsulaiman, Mansour},
journal={IEEE transactions on industrial informatics},
volume={19},
number={2},
pages={2249--2258},
year={2022},
publisher={IEEE}
}

@article{liu2024cross,
title={A cross-session motor imagery classification method based on {R}iemannian geometry and deep domain adaptation},
author={Liu, Wenchao and Guo, Changjiang and Gao, Chang},
journal={Expert Systems with Applications},
volume={237},
pages={121612},
year={2024},
publisher={Elsevier}
}

@article{zhong2024eeg,
  title={EEG-DG: A multi-source domain generalization framework for motor imagery EEG classification},
  author={Zhong, Xiao-Cong and Wang, Qisong and Liu, Dan and Chen, Zhihuang and Liao, Jing-Xiao and Sun, Jinwei and Zhang, Yudong and Fan, Feng-Lei},
  journal={IEEE Journal of Biomedical and Health Informatics},
  volume={29},
  number={4},
  pages={2484--2495},
  year={2024},
  publisher={IEEE}
}

@inproceedings{lotey2022cross,
title={Cross-session motor imagery {EEG} classification using self-supervised contrastive learning},
author={Lotey, Taveena and Keserwani, Prateek and Wasnik, Gaurav and Roy, Partha Pratim},
booktitle={2022 26th International Conference on Pattern Recognition (ICPR)},
pages={975--981},
year={2022},
organization={IEEE}
}

@article{tangermann2012review,
title={Review of the {BCI} Competition {IV}},
author={Tangermann, Michael and M{"u}ller, Klaus-Robert and Aertsen, Ad and Birbaumer, Niels and Braun, Christoph and Brunner, Clemens and Leeb, Robert and Mehring, Carsten and Miller, Kai J and Mueller-Putz, Gernot and others},
journal={Frontiers in Neuroscience},
volume={6},
pages={55},
year={2012},
publisher={Frontiers Media SA}
}

@article{roy2019deep,
title={Deep learning-based electroencephalography analysis: a systematic review},
author={Roy, Yannick and Banville, Hubert and Albuquerque, Isabela and Gramfort, Alexandre and Falk, Tiago H and Faubert, Jocelyn},
journal={Journal of neural engineering},
volume={16},
number={5},
pages={051001},
year={2019},
publisher={IOP Publishing}
}

@techreport{brunner2008bci,
title={{BCI} Competition 2008--{G}raz data set {B}},
author={Brunner, Clemens and Leeb, Robert and M{"u}ller-Putz, Gernot and Schl{"o}gl, Alois and Pfurtscheller, Gert},
institution={Graz University of Technology},
year={2008}
}

@inproceedings{woo2018cbam,
title={{CBAM}: Convolutional block attention module},
author={Woo, Sanghyun and Park, Jongchan and Lee, Joon-Young and Kweon, In So},
booktitle={Proceedings of the European conference on computer vision (ECCV)},
pages={3--19},
year={2018}
}

@inproceedings{ang2008filter,
title={Filter bank common spatial pattern ({FBCSP}) in brain-computer interface},
author={Ang, Kai Keng and Chin, Zheng Yang and Zhang, Haihong and Guan, Cuntai},
booktitle={2008 IEEE international joint conference on neural networks (IEEE world congress on computational intelligence)},
pages={2390--2397},
year={2008},
organization={IEEE}
}

@article{zhao2020deep,
  title={Deep representation-based domain adaptation for nonstationary EEG classification},
  author={Zhao, He and Zheng, Qingqing and Ma, Kai and Li, Huiqi and Zheng, Yefeng},
  journal={IEEE Transactions on Neural Networks and Learning Systems},
  volume={32},
  number={2},
  pages={535--545},
  year={2020},
  publisher={IEEE}
}

@article{lecun2015deep,
title={Deep learning},
author={LeCun, Yann and Bengio, Yoshua and Hinton, Geoffrey},
journal={nature},
volume={521},
number={7553},
pages={436--444},
year={2015},
publisher={Nature Publishing Group UK London}
}

@article{cohen1960coefficient,
title={A coefficient of agreement for nominal scales},
author={Cohen, Jacob},
journal={Educational and psychological measurement},
volume={20},
number={1},
pages={37--46},
year={1960},
publisher={Sage Publications Sage CA: Thousand Oaks, CA}
}

@article{pfurtscheller1999event,
title={Event-related {EEG}/{MEG} synchronization and desynchronization: basic principles},
author={Pfurtscheller, Gert and Da Silva, FH Lopes},
journal={Clinical neurophysiology},
volume={110},
number={11},
pages={1842--1857},
year={1999},
publisher={Elsevier}
}

@inproceedings{ingolfsson2020eeg,
  title={EEG-TCNet: An accurate temporal convolutional network for embedded motor-imagery brain--machine interfaces},
  author={Ingolfsson, Thorir Mar and Hersche, Michael and Wang, Xiaying and Kobayashi, Nobuaki and Cavigelli, Lukas and Benini, Luca},
  booktitle={2020 IEEE international conference on systems, man, and cybernetics (SMC)},
  pages={2958--2965},
  year={2020},
  organization={IEEE}
}

@article{musallam2021electroencephalography,
  title={Electroencephalography-based motor imagery classification using temporal convolutional network fusion},
  author={Musallam, Yazeed K and AlFassam, Nasser I and Muhammad, Ghulam and Amin, Syed Umar and Alsulaiman, Mansour and Abdul, Wadood and Altaheri, Hamdi and Bencherif, Mohamed A and Algabri, Mohammed},
  journal={Biomedical Signal Processing and Control},
  volume={69},
  pages={102826},
  year={2021},
  publisher={Elsevier}
}

@article{kang2009composite,
  title={Composite common spatial pattern for subject-to-subject transfer},
  author={Kang, Hyohyeong and Nam, Yunjun and Choi, Seungjin},
  journal={IEEE Signal Processing Letters},
  volume={16},
  number={8},
  pages={683--686},
  year={2009},
  publisher={IEEE}
}

@article{song2023global,
  title={Global adaptive transformer for cross-subject enhanced EEG classification},
  author={Song, Yonghao and Zheng, Qingqing and Wang, Qiong and Gao, Xiaorong and Heng, Pheng-Ann},
  journal={IEEE Transactions on Neural Systems and Rehabilitation Engineering},
  volume={31},
  pages={2767--2777},
  year={2023},
  publisher={IEEE}
}
 
\end{document}